\documentclass[sigconf]{acmart}

\usepackage{booktabs}
\usepackage{multirow}
\usepackage{xcolor}
\usepackage{amsmath}
\usepackage{enumitem}
\usepackage{tikz}
\usetikzlibrary{arrows.meta,positioning,calc,fit,backgrounds,shapes.geometric}
\usepackage{graphicx}
\usepackage{wrapfig}
\usepackage{algorithm}
\usepackage{algpseudocode}
\usepackage{stfloats}   % allow full-width (table*/figure*) floats at page bottom

\newcommand{\todo}[1]{}
\newcommand{\sd}[1]{\,\textcolor{black!58}{\textmd{\fontsize{8}{8}\selectfont$\pm$#1}}}
\setcopyright{rightsretained}
\renewcommand\footnotetextcopyrightpermission[1]{}

\begin{document}

\title[{Oracle-Budgeted Molecular Optimization with Short-Term Graph Memory}]{Oracle-Budgeted Molecular Optimization \texorpdfstring{\\}{}with Short-Term Graph Memory}

\author{Jiannan Yang}
\affiliation{%
  \department{Computer Science}
  \institution{Stony Brook University}
  \city{Stony Brook}
  \state{New York}
  \country{USA}
}

\author{Veronika Thost}
\affiliation{%
  \department{Molecular AI}
  \institution{Novo Nordisk}
  \city{Lexington}
  \state{Massachusetts}
  \country{USA}
}

\author{Xiang Ling}
\affiliation{%
  \department{Biomedical Engineering}
  \institution{Stony Brook University}
  \city{Stony Brook}
  \state{New York}
  \country{USA}
}

\author{Tengfei Ma}
\affiliation{%
  \department{Biomedical Informatics}
  \institution{Stony Brook University}
  \city{Stony Brook}
  \state{New York}
  \country{USA}
}

\begin{abstract}
Molecular optimization is commonly performed under a limited oracle budget, which makes deciding what to evaluate as important as deciding what to generate. We introduce short-term graph memory, a plug-in module that preserves the generator architecture and native update rule while learning from previously evaluated molecules to prioritize subsequent oracle queries. The module maintains an online graph neural surrogate that pre-screens each round's candidate pool, so the fixed oracle budget is spent on molecules with higher predicted utility. Applied to a fragment-based generator on a standard molecular optimization benchmark, it improves the mean top-10 score at no extra oracle cost and never falls behind the base on any oracle; the gain extends to all four generators we tested at a tight budget of one thousand calls. We then analyze how surrogate-guided selection interacts with the exploration and exploitation behavior of different generators. Its benefit at larger budgets is consistent with two properties of the backbone: how broadly it searches, and how effectively its native search already exploits oracle feedback. We provide a simple way to spend a fixed oracle budget more selectively, and evidence on which generators benefit from it.
\end{abstract}

\keywords{molecular optimization, oracle-efficient search, drug discovery}

\maketitle
\renewcommand{\thefootnote}{}%
\footnotetext{Code: \url{https://github.com/JPaulYang/short-term-graph-memory}}%
\renewcommand{\thefootnote}{\arabic{footnote}}%
\pagestyle{plain}   % arXiv preprint: drop the acmart "Conference'17" running head

\begin{figure}[t]
\centering
\includegraphics[width=0.92\columnwidth]{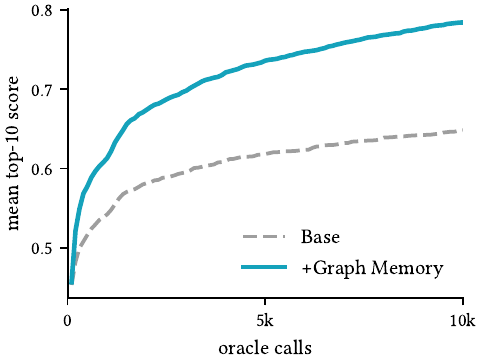}
\caption{Mean top-10 score vs.\ oracle calls on GenMol, averaged over the benchmark
oracles. Short-term graph memory (solid) reaches high-scoring molecules with far
fewer oracle calls than the base generator (dashed), at no extra oracle cost.}
\Description{Line plot of mean top-10 score against number of oracle calls on GenMol. The
solid curve for short-term graph memory rises more steeply and stays above the dashed curve
for the base generator across the whole budget.}
\label{fig:climbing}
\end{figure}

%======================================================================
\section{Introduction}
\label{sec:intro}

\begin{figure*}[t]
\centering
\begin{tikzpicture}[font=\normalsize,
  box/.style={draw,rounded corners,minimum height=9.5mm,align=center,inner sep=4pt},
  molhi/.style={draw=green!55!black,line width=1.1pt,rounded corners=1pt,inner sep=0.8pt,fill=white},
  mollo/.style={draw=black!30,line width=0.8pt,rounded corners=1pt,inner sep=0.8pt,fill=white},
  seed/.style={draw=blue!45,line width=1pt,rounded corners=1pt,inner sep=0.8pt,fill=blue!4},
  mem/.style={draw=green!45!black,rounded corners=2pt,fill=green!7},
  buf/.style={cylinder,shape border rotate=90,aspect=0.22,draw=black!45,fill=black!5,
              minimum width=18mm,minimum height=8mm,inner sep=1.5pt,font=\small,align=center},
  arr/.style={-{Latex[length=2mm]},semithick},
  thin2/.style={-{Latex[length=1.5mm]},black!30},
  warr/.style={-{Latex[length=2.6mm]},line width=1.2pt,draw=green!45!black},
  fb/.style={-{Latex[length=2mm]},black!55,semithick},
  gfb/.style={-{Latex[length=2mm]},green!45!black,semithick},
  note/.style={font=\small,align=center}]

  % ===================== (a) Baseline =====================
  \node[seed] (xA) {\includegraphics[width=9.5mm]{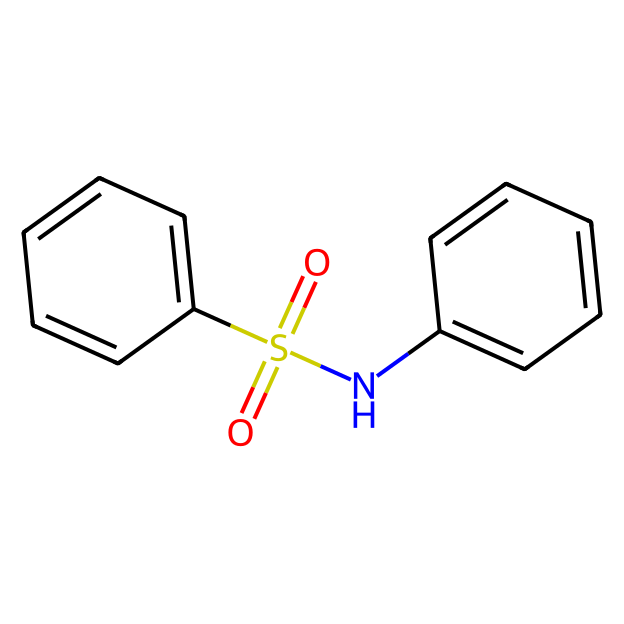}};
  \node[note,above=0.4mm of xA,text=blue!40!black] {seed $x$};
  \node[box,fill=blue!8,right=5mm of xA] (gA) {Generator};
  \node[mollo] (a1) at ($(gA.east)+(18mm,11.5mm)$)  {\includegraphics[width=9.5mm]{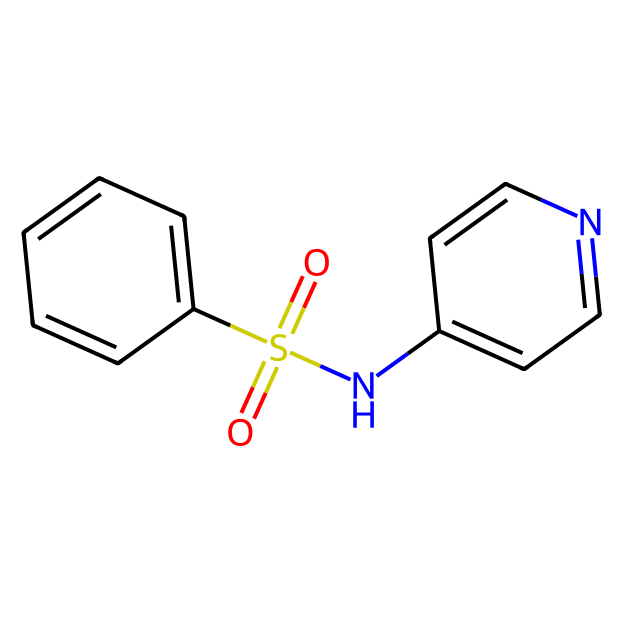}};
  \node[mollo] (a2) at ($(gA.east)+(18mm,0mm)$)     {\includegraphics[width=9.5mm]{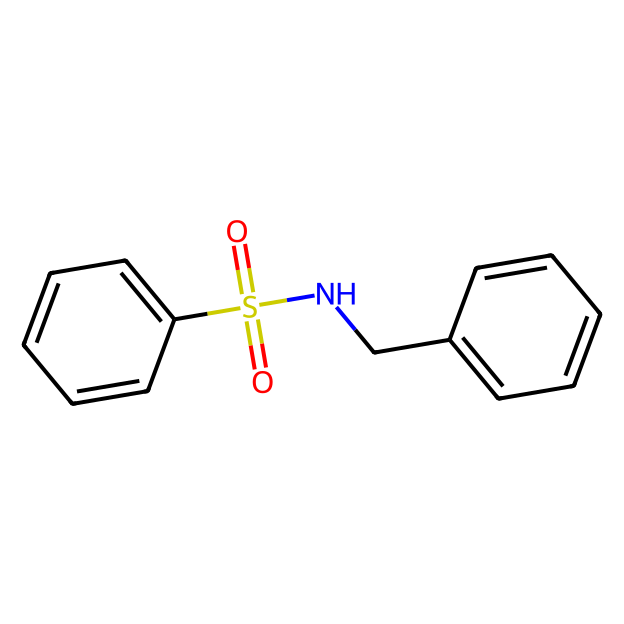}};
  \node[mollo] (a3) at ($(gA.east)+(18mm,-11.5mm)$) {\includegraphics[width=9.5mm]{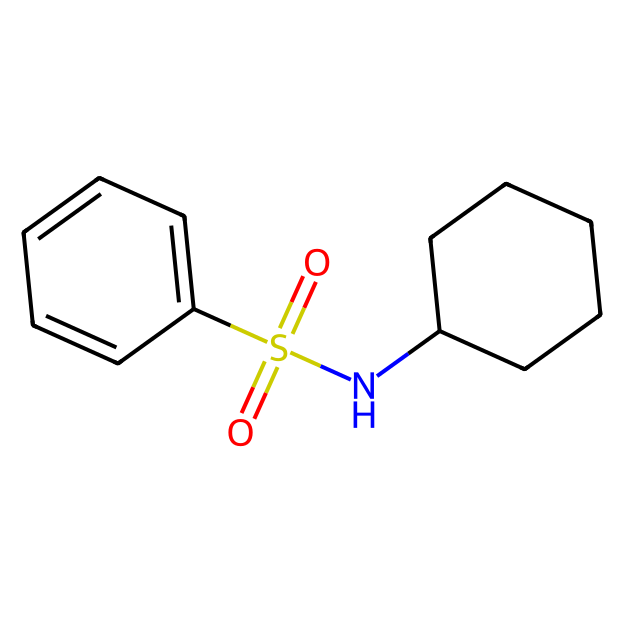}};
  \node[box,fill=red!8] (oA) at ($(gA.east)+(41mm,0)$) {Oracle $f$};
  \node[buf] (cA) at ($(oA.south)+(0,-16mm)$) {$(x,f(x))$};
  \draw[arr] (xA) -- (gA);
  \draw[arr] (gA) -- node[above,pos=0.42,font=\small]{$N$ cand.} (a2.west);
  \foreach \i in {a1,a2,a3}{\draw[thin2] (\i.east) -- (oA.west);}
  \draw[fb] (oA.south) -- (cA.north);
  \draw[fb] (cA.south) -- ++(0,-6mm) -| (xA.south);
  \node[note,text=black!55] at ($(cA.south)+(-17mm,-8mm)$) {keep best $x$};
  \node[note,text=red!60,align=center,above=2.2mm of oA] {budget spread over\\ weak candidates};
  \node[font=\bfseries] at ($(xA.south)!0.5!(oA.south)+(0,-30mm)$) {(a) Baseline};

  % ================= (b) + Graph memory =================
  \node[seed] (xB) at ($(oA.east)+(18mm,0)$) {\includegraphics[width=9.5mm]{figs/mols/n_seed.png}};
  \node[note,above=0.4mm of xB,text=blue!40!black] {seed $x$};
  \node[box,fill=blue!8,right=5mm of xB] (gB) {Generator};
  \node[molhi] (b1) at ($(gB.east)+(19mm,11.5mm)$)  {\includegraphics[width=9.5mm]{figs/mols/n_hi1.png}};
  \node[molhi] (b2) at ($(gB.east)+(19mm,0mm)$)     {\includegraphics[width=9.5mm]{figs/mols/n_hi2.png}};
  \node[mollo] (b3) at ($(gB.east)+(19mm,-11.5mm)$) {\includegraphics[width=9.5mm]{figs/mols/n_lo.png}};
  \begin{scope}[on background layer]
    \node[mem,fit=(b1)(b2)(b3),inner sep=2mm] (memB) {};
  \end{scope}
  \node[note,text=green!35!black,anchor=south] at ($(memB.north)+(0,-0.6mm)$) {graph memory $s_\theta$};
  \node[box,fill=red!8] (oB) at ($(gB.east)+(47mm,0)$) {Oracle $f$};
  \node[buf] (cB) at ($(oB.south)+(0,-16mm)$) {$(x,f(x))$};
  \draw[arr] (xB) -- (gB);
  \draw[arr] (gB) -- node[above,pos=0.42,font=\small]{$N$ cand.} (memB.west);
  \draw[warr] (memB.east) -- node[above,text=green!45!black]{top-$k$} (oB.west);
  \draw[fb] (oB.south) -- (cB.north);
  \draw[fb] (cB.south) -- ++(0,-6mm) -| (xB.south);
  \node[note,text=black!55] at ($(cB.south)+(-21mm,-8mm)$) {keep best $x$};
  % green feedback: from cylinder bottom (~1/4 from left) down, left, then up into memory
  \coordinate (gtop) at ($(cB.south)+(-4mm,0)$);
  \coordinate (gr)   at ($(gtop)+(0,-3mm)$);
  \coordinate (gl)   at (gr -| memB.south);
  \draw[gfb] (gtop) -- (gr) -- (gl) -- (memB.south);
  \node[note,text=green!40!black,above=0.4mm] at ($(gr)!0.5!(gl)$) {online fine-tune};
  \node[note,text=green!40!black,align=center,above=2.2mm of oB] {budget on\\ best candidates};
  \node[font=\bfseries] at ($(xB.south)!0.5!(oB.south)+(0,-30mm)$) {(b) + Graph memory};

  % ---- divider ----
  \draw[dashed,black!25] ($(oA.east)!0.5!(xB.west)+(0,-34mm)$) -- ++(0,52mm);
\end{tikzpicture}
\caption{A seed-based optimization loop under the same generator and oracle budget.
\textbf{(a)} The base generator evaluates proposals with no learned pre-screening. \textbf{(b)} Graph
memory $s_\theta$, fine-tuned online on the same evaluations, pre-screens them so only the
selected ones are evaluated.}
\Description{Two-panel schematic of one optimization round. In panel (a) the generator turns
a seed molecule into several candidates, all of which are sent to the oracle, and the best
scoring one becomes the next seed. In panel (b) the same candidates first pass through a
graph memory that ranks them, only the top ones reach the oracle, and the resulting
evaluations both update the seed and fine-tune the memory.}
\label{fig:compare}
\end{figure*}

Molecular optimization is often limited not by the ability to generate candidates,
but by the cost of evaluating them. Modern generative models can propose large
numbers of molecules at relatively low computational cost, whereas evaluating their
desired properties may require expensive simulations, docking procedures, learned
proxy oracles, or experimental validation. This asymmetry makes the oracle, not the generator, the bottleneck: a method can propose far
more molecules than it can ever afford to test, so its progress is governed by the evaluations
it can make rather than by the candidates it can produce.

That cost also bounds how many evaluations a campaign can afford at all, and in the settings
these oracles ultimately model the number is very small. A wet-lab campaign may synthesize and assay only a handful to a few dozen
molecules~\cite{zhavoronkov2019ddr1}, and even efforts that screen enormous virtual libraries
are triaged down to a few dozen experimental tests~\cite{stokes2020antibiotic,scalia2025virtual}.
A thousand true evaluations already sit at the upper end of what an academic or early-stage
effort can afford, and tens of thousands are rarely realistic. Progress therefore hinges on
the tight-budget regime, where every oracle call must be spent well.

Recent molecular generation methods achieve strong performance on oracle-budgeted
benchmarks~\cite{pmo} by combining stochastic candidate generation with chemically
informed priors over which candidates to propose, and a growing body of work makes the
expensive evaluation the primary cost to reduce, through sample-efficient
search policies and learned surrogates~\cite{kim2024geneticguided,lee2025nfbo,nguyen2025lico}.
These optimizers use oracle feedback in different ways to shape future proposals, but they
may still evaluate many weak candidates before identifying high-value regions. Under a fixed
budget, how efficiently good molecules are found depends not on generation alone but on a
complementary allocation problem: from each generated pool, which candidates should consume
the next oracle calls?

In this work, we study whether previously evaluated molecules can be recycled as a
short-term memory for future oracle allocation, and whether a common mechanism can serve
optimizers whose search dynamics differ. We propose a plug-in \emph{short-term graph memory}
module that continuously fine-tunes a graph neural surrogate on molecules already evaluated by
the oracle, and then uses it as a cheap structure-aware selector between the generator and the
oracle. Because the module preserves each backbone's architecture and native update rule, the
same interface can be applied across fragment-based, flow-based, policy-gradient, and genetic
search. The budget counts only oracle evaluations, so the module is
\emph{oracle-budget-neutral}: it reallocates the expensive budget rather than enlarging it. It
does add cheap computation, since it draws a larger candidate pool and runs surrogate
inference and online training on it.

Empirically, short-term graph memory yields large gains on generators that propose a
broad, weakly-ranked candidate pool: for GenMol~\cite{genmol} initialized from its
full unscored fragment pool over the practical molecular optimization (PMO)
benchmark~\cite{pmo}, it raises the mean top-10 score from
$0.648$ to $0.784$ and is no worse than the base generator on any of the 22 oracles, using
\emph{no additional oracle calls} (Section~\ref{sec:results}); it also improves the
discrete-flow generator InVirtuoGen~\cite{invirtuogen}, and, with a diversity-preserving
selector, REINVENT~\cite{olivecrona2017reinvent}. At the tight budget of $1{,}000$
calls, the more practically feasible of the two regimes, memory improves the mean top-10 score
of all four generators we tested. Its benefit at larger budgets varies with the base
generator's search dynamics: a genetic algorithm that already concentrates its own search
shows little net change. Selection rule matters too. Hard filtering destabilizes REINVENT,
because the selected candidates are also its policy-gradient training batch
(Appendix~\ref{app:reinvent}). We analyze this generator-memory compatibility and the
exploration-exploitation trade-off that memory induces.

\paragraph{Contributions.}
\begin{enumerate}[leftmargin=*]
  \item We introduce \textbf{short-term graph memory}, an oracle-budget-neutral module that
  recycles already-evaluated molecules into an online graph surrogate and re-ranks each
  round's candidate pool, leaving the backbone's architecture and native update rule
  untouched (Section~\ref{sec:method}).
  \item We show that one such selector improves all four backbones at a tight budget and
  without additional oracle calls, spanning masked-diffusion, discrete-flow,
  policy-gradient, and genetic search (Section~\ref{sec:results}).
  \item We characterize where the gain is retained and where it diminishes as the budget grows,
  tying it to how far a backbone already exploits its own feedback, show that the exploitation pressure behind it
  is adjustable rather than fixed, and compare against Augmented Memory~\cite{augmem} on a
  shared backbone (Section~\ref{sec:analysis},~\ref{sec:augmem}).
\end{enumerate}

%======================================================================
\section{Problem Formulation}
\label{sec:problem}

We consider sequential molecular optimization under a fixed oracle budget. At round $t$,
a generator $G$ proposes a candidate pool $P_t \subseteq \mathcal{X}$ from the evaluation
history $\mathcal{M}_{t-1} = \{(x_i, f(x_i))\}_{i=1}^{n_{t-1}}$. A selector chooses a
subset $S_t \subseteq P_t$ to be scored by an expensive oracle
$f:\mathcal{X}\to[0,1]$, subject to
\[
  \sum_t |S_t| \;\le\; B .
\]
The evaluated pairs are appended to the history
$$\mathcal{M}_t = \mathcal{M}_{t-1}
\cup \{(x,f(x)) : x \in S_t\}$$
which both the generator and the selector may use in
the next round.

Candidate generation and surrogate inference are treated as unbudgeted computation:
$B$ counts oracle evaluations only. This follows the accounting convention of the PMO
benchmark~\cite{pmo}, and it reflects settings where a single evaluation is a docking
run, an assay, or a synthesis, and is orders of magnitude more expensive than sampling
a molecule. Under this convention the design question is not how to generate more
molecules, but how to allocate the $B$ evaluations across the molecules already
proposed.

%======================================================================
\section{Short-Term Graph Memory}
\label{sec:method}

\subsection{Framework and backbone interface}
Molecular optimizers are feedback-driven, but many use that feedback narrowly.
GenMol~\cite{genmol} and InVirtuoGen~\cite{invirtuogen} build molecules from fragments and
reuse feedback only through which evaluated molecules seed the next round.
Graph-GA~\cite{jensen2019graphga} applies fixed crossover and mutation operators and reuses
feedback through fitness-weighted selection of the population they act on. In these three the
operator's parameters never change. REINVENT~\cite{olivecrona2017reinvent} differs: feedback
enters as reward and updates the generator's weights directly, which is also why an external
selector interacts differently with it. None of them, however, ranks the candidates it
proposes within a round. Short-term graph memory addresses that gap: it adds a learned
selector over each round's proposal pool, without altering the backbone architecture or its
native update rule.

Short-term graph memory is a plug-in module inserted between the generator and the oracle, summarized in
Algorithm~\ref{alg:graphmem}. Each round the generator proposes a pool, the surrogate
scores it cheaply, only the selected candidates are sent to the oracle, and the new
evaluations both accumulate in the buffer $\mathcal{M}$ and fine-tune the surrogate online.
The module does not increase the oracle budget; it changes which molecules the same budget
is spent on. Figure~\ref{fig:compare} contrasts the two allocations.

The module needs only two things from a backbone: (1) a stream of candidate
SMILES~\cite{weininger1988smiles}
and (2) a scalar oracle. Everything else stays with the generator, including the call
$G.\textsc{update}(\mathcal{M})$ in Algorithm~\ref{alg:graphmem}, which stands for the native
rule each backbone already applies to accumulated feedback: pool reseeding, population
selection, or a policy-gradient step. The memory replaces none of these; it inserts a
ranking stage immediately before the oracle call. The interface is uniform across backbones,
while what it wraps is not.

\subsection{Online graph surrogate}
\label{subsec:surrogate}
The surrogate is a graph neural network $s_\theta$ over the 2D molecular graph. We use a
GraphGPS backbone~\cite{graphgps} that combines GINE~\cite{xu2018how,hu2020strategies}
message passing with attention~\cite{vaswani2017attention} and positional encodings, mean-pools
node embeddings, and maps them to a scalar through a linear head. The backbone starts from a checkpoint pretrained on ZINC~\cite{sterling2015zinc} with the motif-prediction objective
of~\cite{yangself}, so the surrogate begins the run with
general chemical structure already encoded and only has to specialize to the current oracle.
Architecture and optimizer settings are listed in Appendix~\ref{app:impl}.

The surrogate is then fine-tuned online. Selection stays random during an initial warm-up of
$w$ evaluated molecules, which avoids ranking by an unadapted model. After that, whenever $u$
new oracle calls have accumulated since the last update, we take a single gradient step on a
mini-batch sampled from the buffer, minimizing
\[
\mathcal{L}(\theta) = \frac{1}{|\mathcal{B}|}\sum_{(x,y)\in\mathcal{B}} \big(s_\theta(x)-y\big)^2 ,
\qquad \mathcal{B}\subseteq\mathcal{M}.
\]
The buffer holds true oracle scores only, so the surrogate never trains on its own
predictions, and the memory is short-term in a strict sense: the checkpoint has seen no
oracle score, so knowledge of the objective accrues only within the run.
The backbone-specific values of $w$ and $u$ are listed in Appendix~\ref{app:impl}.

\begin{algorithm}[t]
\caption{Short-term graph memory}
\label{alg:graphmem}
\begin{algorithmic}[1]
\Require generator $G$, oracle $f$, budget $B$, surrogate $s_\theta$, pool size $N$, forward count $k$, warm-up $w$, update period $u$
\State $\mathcal{M} \gets \emptyset$;\ \ $\ell \gets 0$
\While{$|\mathcal{M}| < B$}
  \State $P \gets G.\textsc{propose}(N)$
  \State $k_t \gets \min(k,\ B - |\mathcal{M}|)$
  \If{$|\mathcal{M}| \ge w$}
    \State $S \gets \textsc{select}_{k_t}\big(P,\ s_\theta(P)\big)$ \Comment{deterministic or stochastic}
  \Else
    \State $S \gets \textsc{random}(P, k_t)$
  \EndIf
  \For{$x \in S$}
    \State $y \gets f(x)$;\ \ $\mathcal{M} \gets \mathcal{M} \cup \{(x,y)\}$
  \EndFor
  \If{$|\mathcal{M}| \ge w$ \textbf{and} $|\mathcal{M}| - \ell \ge u$}
    \State $s_\theta \gets \textsc{fine-tune}(s_\theta,\ \mathcal{M})$;\ \ $\ell \gets |\mathcal{M}|$
  \EndIf
  \State $G.\textsc{update}(\mathcal{M})$ \Comment{backbone-specific}
\EndWhile
\State \Return top molecules in $\mathcal{M}$
\end{algorithmic}
\end{algorithm}

\subsection{Candidate selection}
\label{subsec:selection}
Given a pool $P$ of $N$ proposals and surrogate scores $s_\theta(P)$, the selector returns
the $k$ molecules to evaluate. We study two rules. The \emph{deterministic} selector takes
the $k$ highest-scored candidates. It maximizes exploitation of the current surrogate and is
our default. The \emph{stochastic} selector instead samples $k$ candidates without
replacement from $\mathrm{softmax}\big(s_\theta(P)/T\big)$ with temperature $T$, so
higher-scored candidates are more likely to be chosen but low-scored ones are not
categorically excluded. Temperature is a knob on exploitation strength, not a new optimizer,
and requires no reinforcement-learning machinery. The two rules differ only in how the same
surrogate scores become selections.

Hard selection can be unstable when the selected subset is also what updates a parametric
generator, as it is in REINVENT: the selector's bias returns through the policy and is
amplified until generation collapses onto duplicates and the run stalls. We therefore use the
stochastic selector for REINVENT and the deterministic one elsewhere.
Appendix~\ref{app:reinvent} gives the mechanism and the runs behind this choice.

%======================================================================
\section{Experimental Setup}
\label{sec:setup}

\subsection{Benchmark and evaluation protocol}
We evaluate on the practical molecular optimization benchmark~\cite{pmo}, a panel of
property oracles that scores optimizers under an explicit query budget rather than by
generation quality alone. The benchmark evaluates top-$k$ score and top-$k$ AUC, and we use $k=10$ for both:
final top-10 score as the measure of solution quality and
top-10 AUC as the measure of climbing speed: how quickly high-scoring molecules are
discovered as the budget is consumed. We use $B=10{,}000$ oracle calls,
the benchmark default, and additionally report every result at $1{,}000$ calls by reading
each run's trajectory at that point. The tight budget is closer to what a real evaluation
campaign can afford, so we treat both as primary rather than treating the tight budget as an
ablation.

The tight-budget results average three random seeds for every backbone, oracle, and
condition; the full-budget results are a single run with a pinned seed.
Appendix~\ref{app:auc} reports the per-oracle standard deviations. We test whether memory
exceeds base with a one-sided Wilcoxon signed-rank test over the oracles, together with the
win/tie/loss count, which measures how consistently the sign of the improvement holds across
tasks. We report aggregates on 22 of the 23
oracles, excluding the valsartan SMARTS task
because its outcomes are near-binary and were highly sensitive to whether the target
substructure was encountered in a given run. Appendix~\ref{app:valsartan} gives the excluded
results and a sensitivity analysis; including this oracle does not change the qualitative
conclusions.

\begin{table*}[t]
\caption{Top-10 score and top-10 AUC for each generator with and without memory, at a tight
($1{,}000$) and a full ($10{,}000$) oracle budget. The $p$-value is a one-sided Wilcoxon
signed-rank test paired across the 22 oracles. Per-oracle results are in
Appendix~\ref{app:auc}.\label{tab:main-final}}
\normalsize
\begin{tabular*}{\textwidth}{@{\extracolsep{\fill}} l ccc ccc ccc ccc @{}}
\toprule
 & \multicolumn{3}{c}{Top-10 @ $1{,}000$} & \multicolumn{3}{c}{Top-10 AUC @ $1{,}000$} & \multicolumn{3}{c}{Top-10 @ $10{,}000$} & \multicolumn{3}{c}{Top-10 AUC @ $10{,}000$} \\
\cmidrule(lr){2-4}\cmidrule(lr){5-7}\cmidrule(lr){8-10}\cmidrule(lr){11-13}
Backbone & Base & +Mem & $p$ & Base & +Mem & $p$ & Base & +Mem & $p$ & Base & +Mem & $p$ \\
\midrule
GenMol & 0.542 & \textbf{0.613} & \textbf{$<$0.001} & 0.537 & \textbf{0.606} & \textbf{$<$0.001} & 0.648 & \textbf{0.784} & \textbf{$<$0.001} & 0.603 & \textbf{0.712} & \textbf{$<$0.001} \\
InVirtuoGen & 0.532 & \textbf{0.569} & \textbf{0.002} & 0.450 & \textbf{0.474} & \textbf{0.049} & 0.687 & \textbf{0.730} & \textbf{0.002} & 0.621 & \textbf{0.668} & \textbf{0.002} \\
REINVENT & 0.531 & \textbf{0.571} & \textbf{$<$0.001} & 0.419 & \textbf{0.453} & \textbf{$<$0.001} & \textbf{0.780} & 0.762 & 0.558 & 0.686 & \textbf{0.696} & \textbf{0.015} \\
Graph-GA & 0.530 & \textbf{0.557} & \textbf{0.009} & 0.439 & \textbf{0.457} & \textbf{$<$0.001} & \textbf{0.753} & 0.733 & 0.410 & 0.652 & \textbf{0.653} & 0.099 \\
\bottomrule
\end{tabular*}
\end{table*}

\subsection{Backbones and experimental configuration}
We instantiate the module on four backbones spanning distinct search families:
GenMol~\cite{genmol}, masked-diffusion fragment recombination; InVirtuoGen~\cite{invirtuogen},
discrete-flow fragment generation; REINVENT~\cite{olivecrona2017reinvent}, a SMILES
recurrent-network policy gradient; and Graph-GA~\cite{jensen2019graphga}, a graph genetic
algorithm. Full per-backbone configurations are in Appendix~\ref{app:impl}. For the two
fragment generators, memory ranks pools of 16 candidates and forwards one per round. For
REINVENT and Graph-GA it filters half of each proposal batch. REINVENT uses stochastic
selection at temperature $T{=}1$, for the reason given in Section~\ref{subsec:selection};
the other three use deterministic selection.

Base and memory runs use the same backbone and the same total oracle budget. Graph memory
adds a pre-oracle selection step, while each backbone retains its native generation and
update mechanism. For InVirtuoGen, REINVENT, and Graph-GA the base run draws the same pool
and selects from it without the surrogate: at random for InVirtuoGen, and by evaluating the
whole pool for REINVENT and Graph-GA. For GenMol the base run follows the original
one-evaluation-per-round loop, so the memory run draws a larger pool than its base. The two
runs always spend an identical number of oracle calls, which is what the budget counts; the
memory run spends more unbudgeted generation and surrogate compute. For REINVENT and Graph-GA,
forwarding only half of each batch also means the memory run completes roughly twice as many
proposal and native-update rounds within the same budget. That is part of what reallocating
the budget does, so these two backbones do not separate surrogate ranking from update
frequency.

Both GenMol runs start from the full unscored fragment pool, whereas the published
configuration seeds recombination from an oracle-ranked top-100 pool. Starting unranked keeps
all candidate ordering in the selector under study, which is what the comparison is about. It
is also the harder starting point, and our GenMol results refer to it throughout.

%======================================================================
\section{Main Results}
\label{sec:results}

Figure~\ref{fig:climbing} shows the effect on GenMol: memory reaches
high-scoring molecules with far fewer oracle calls. Table~\ref{tab:main-final} reports
every generator with and without memory at a tight and a full oracle budget, and
Appendix~\ref{app:auc} breaks every cell of it down by oracle.

GenMol shows the largest effect in Table~\ref{tab:main-final}, at both budgets and on both
metrics, and at no extra oracle cost. That average hides nothing: per oracle, memory is 
better than the base on almost all of them
(Appendix~\ref{app:auc}). The gain there concentrates on the hard single-target objectives,
where the base proposes capable molecules but scores too many weak ones to surface them.

At the tight budget memory improves the mean top-10 score of all four backbones, with
positive paired differences across the oracle panel in each case. This is the regime the
benchmark is built to stress, and the closer of the two to what an evaluation campaign can
afford, so the effect is not confined to a single generator family.

What changes with budget is how long the lead lasts. By ten thousand calls the two adaptive
searchers have caught up on final score: REINVENT and Graph-GA each match or exceed their
memory run at the finish, while GenMol and InVirtuoGen keep their advantage. Climbing speed
tells a different story from final score here. Top-10 AUC stays ahead for three of the four
backbones at the full budget, and is comparable between the two runs on Graph-GA, so memory
reaches good molecules sooner even where it does not end higher.

These budget-dependent patterns suggest that memory interacts with both the breadth of a
generator's proposals and the way the generator already uses its own feedback. We analyze
that next.

\begin{figure*}[t]
\centering
\includegraphics{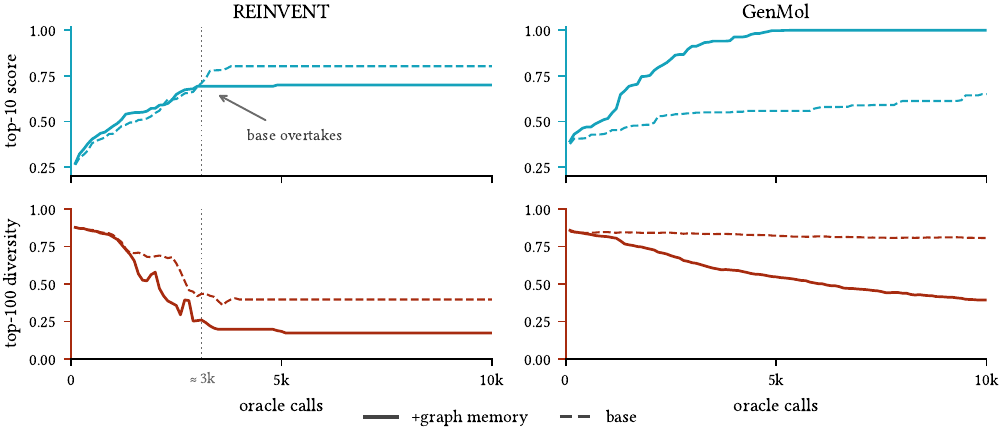}
\caption{\textbf{REINVENT and GenMol on mestranol similarity.} Top-10 score (top, blue)
and top-100 diversity (bottom, red), with memory (solid) and without (dashed). The dotted
line marks where base overtakes memory on REINVENT.}
\Description{Four line plots arranged in two rows and two columns, REINVENT on the left and
GenMol on the right. The top row plots top-10 score against oracle calls, the bottom row
top-100 diversity. On REINVENT the memory curve leads early but is overtaken by the base
generator after roughly three thousand calls, while on GenMol it stays above the base
throughout. Diversity declines under memory for both backbones, further on REINVENT.}
\label{fig:crossover}
\end{figure*}

%======================================================================
\section{Generator--Memory Compatibility}
\label{sec:analysis}
Memory improved all four backbones at the tight budget; at the full budget only two kept a
score advantage. We ask first what separates them, then what the gain costs.

\subsection{Conditions for improvement}

\begin{table}[t]
\caption{How each backbone natively uses oracle feedback. Per-backbone pool sizes and selector
settings are in Table~\ref{tab:setup}.}
\label{tab:compat}
\normalsize
\begin{tabular*}{\columnwidth}{@{\extracolsep{\fill}} l l l @{}}
\toprule
Backbone & Proposes over & Feedback re-enters via \\
\midrule
GenMol      & \textsc{safe} strings~\cite{noutahi2024gotta}   & seeds, drawn unweighted \\
InVirtuoGen & fragment \textsc{smiles} & seeds, score-weighted \\
Graph-GA    & molecular graphs        & seeds, fitness-weighted \\
REINVENT    & \textsc{smiles}         & generator weights \\
\bottomrule
\end{tabular*}
\end{table}

Memory adds value when its pre-oracle ranking provides information beyond the backbone's
native use of feedback. Table~\ref{tab:compat} orders the four backbones by how strongly that
native use concentrates their future proposals. GenMol draws its seeds unweighted, so feedback
shapes what it proposes least; InVirtuoGen and Graph-GA weight the seeds by score, and
REINVENT folds feedback into the generator's own weights. At the tight budget memory helps all
four, because the external surrogate supplies a signal the native search has not yet
acquired, which is where the gain in Table~\ref{tab:main-final} comes from. As the budget
grows, the backbones that concentrate most strongly acquire that signal themselves and the
marginal benefit of memory decreases; the advantage persists for GenMol and InVirtuoGen.

\subsection{Exploitation and diversity}
\label{sec:variant}
Memory is useful because it is selective: it repeatedly favors candidates that the surrogate
associates with previously observed high scores. In our runs this selectivity shows up as
concentration in chemical space. We quantify it as top-100 diversity, one minus the
mean pairwise Tanimoto similarity over 2048-bit Morgan fingerprints of radius
2~\cite{rogers2010extended,tanimoto1958elementary}. This is the one place we depart from
$k{=}10$, because ten molecules give only 45 pairs and are easily dominated by a single
scaffold even when the wider search stays diverse. Table~\ref{tab:diversity} shows the
effect on GenMol. Where memory drives a large score gain, the discovered molecules
concentrate on a few scaffolds; where the gain is modest, diversity is largely preserved.
The score gain is bought with exploitation.

\begin{table}[H]
\caption{Top-100 diversity on GenMol for three oracles. Deterministic memory reduces it most
where the score gain is largest.}
\label{tab:diversity}
\normalsize
\begin{tabular*}{\columnwidth}{@{\extracolsep{\fill}} l ccc @{}}
\toprule
Oracle & Base & Memory (det.) & Memory (stoch.) \\
\midrule
jnk3       & 0.807 & 0.296 & 0.572 \\
gsk3b      & 0.815 & 0.252 & 0.620 \\
amlodipine & 0.784 & 0.668 & 0.760 \\
\bottomrule
\end{tabular*}
\end{table}

Whether that exploitation keeps paying off appears to depend on how broadly the base
generator searches. We use the
top-100 diversity of the unguided base generator as an empirical proxy for search breadth.
Averaged over the panel at the full budget it is $0.82$ for GenMol, $0.58$ for Graph-GA, and
$0.42$ for REINVENT, so the backbones differ substantially on this axis before any memory is
added. Figure~\ref{fig:crossover} contrasts the two ends of that range on one oracle. With REINVENT,
memory climbs fast and leads for thousands of calls, but drives diversity below even the base
generator's, and after the search narrows the base overtakes it. With GenMol the base stays
broadly diverse throughout; memory still concentrates the search, yet stays above base at
every budget.

These observations are consistent with search breadth moderating the effect of
memory-induced exploitation. We state this as a hypothesis rather than an established rule:
it rests on one oracle in Figure~\ref{fig:crossover} and three in
Table~\ref{tab:diversity}, and breadth is measured only through a diversity proxy.

Whatever moderates the effect, the pressure itself should be adjustable: if the gain is
bought with exploitation, relaxing the selection rule should trade score back for
diversity. The stochastic selector of Section~\ref{subsec:selection} tests this
directly, since it changes only how the same surrogate scores become selections.

\begin{figure}[t]
\centering
\includegraphics[width=\columnwidth]{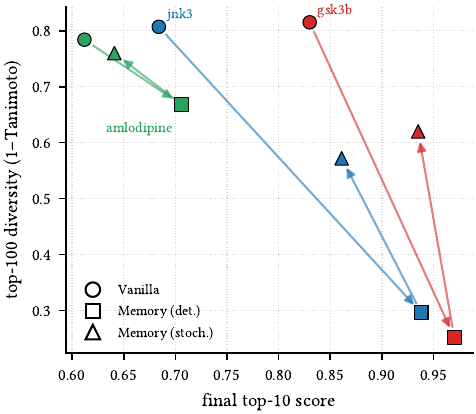}
\caption{\textbf{Score and diversity on GenMol.} Points trace each oracle from the base
generator (circle) to deterministic memory (square) to stochastic memory (triangle).}
\Description{Scatter plot of top-100 diversity against final top-10 score on GenMol. Each
oracle contributes three connected markers: a circle for the base generator, a square for
deterministic memory at higher score and lower diversity, and a triangle for stochastic
memory lying between the two.}
\label{fig:tradeoff}
\end{figure}

The stochastic selector recovers much of the lost diversity, as
Table~\ref{tab:diversity} and Figure~\ref{fig:tradeoff} show, though its best score is
generally at or below deterministic filtering. This supports reading the
score gain as exploitation pressure that can be relaxed in a controllable way. The
stochastic selector complements deterministic memory when diversity is wanted, rather than
replacing it.

%======================================================================
\section{Comparison with Augmented Memory}
\label{sec:augmem}

Short-term graph memory recycles oracle feedback into a selector. An alternative is to
recycle it into the generator. Augmented Memory~\cite{augmem} is the closest method of that
kind: it buffers the highest-scoring molecules and replays them as randomized SMILES for
extra policy-gradient updates, so the feedback reshapes what the generator proposes.
One approach edits the generator's training loop, the other adds a selector in front of it.
Appendix~\ref{app:augmem} gives the algorithm.

We compare the two injection points on GenMol under the same $10{,}000$-call budget. Both
improve over the base, so the gain is not unique to external re-ranking. Injecting it into
the selector is the stronger of the two here, winning on 16 of the 22 oracles and retaining
more top-100 diversity. Repeatedly regenerating around the same buffered molecules pulls the
population toward a few scaffolds, whereas re-ranking leaves the proposal operator untouched.
Appendix~\ref{app:augmem} details the GenMol adaptation.

\begin{table}[t]
\caption{Augmented Memory~\cite{augmem}, adapted to GenMol as described in
Appendix~\ref{app:augmem}, vs.\ graph memory at a $10{,}000$-call budget.\label{tab:augmem}}
\normalsize
\begin{tabular*}{\columnwidth}{@{\extracolsep{\fill}} l ccc @{}}
\toprule
GenMol & Base & \shortstack{$+$Augmented\\Memory} & \shortstack{$+$Graph\\memory (ours)} \\
\midrule
Mean top-10 score        & 0.648 & 0.744 & \textbf{0.784} \\
Mean top-100 diversity   & 0.821 & 0.611 & \textbf{0.637} \\
Wins/ties/losses & -- & -- & 16/4/2 \\
\bottomrule
\end{tabular*}
\end{table}

%======================================================================
\section{Related Work}
\label{sec:related}

\paragraph{Search efficiency under a fixed budget.}
Oracle-budgeted molecular optimization has shifted from raw generation quality toward
search efficiency under a limited evaluation budget, standardized by the PMO
benchmark~\cite{pmo}, which shows that many nominal state-of-the-art generators lose
their advantage once query efficiency is enforced. Recent methods treat expensive
evaluation as the bottleneck and design a learned search procedure around it:
genetic-guided GFlowNets distill a genetic algorithm into a GFlowNet
policy~\cite{kim2024geneticguided}, NF-BO replaces variational-autoencoder latent Bayesian
optimization with autoregressive normalizing flows~\cite{lee2025nfbo}, and LICO adapts
a large language model as an in-context surrogate, especially in the low-budget
regime~\cite{nguyen2025lico}. These strengthen the generator or the search policy;
short-term graph memory instead strengthens the \emph{selector} between the generator and
the oracle, leaving the generator's architecture and update rule as they are.

\paragraph{Surrogate-assisted candidate selection.}
Fitting a predictor online on already-evaluated molecules and using it to pre-rank
dynamically generated candidates is an established mechanism: GP-BO fits a Gaussian process
within the run and uses an acquisition function to choose which of a genetic algorithm's
offspring to evaluate, and is among the strongest methods on the PMO
benchmark~\cite{tripp2021fresh,pmo}. Rather than build a stronger critic for one
evolutionary framework, we ask whether a single external graph selector can augment
optimizers whose native search dynamics differ. Our selector is correspondingly simple: a
GraphGPS~\cite{graphgps} predictor fine-tuned online, ranking by predicted score without
uncertainty estimates or an acquisition function.

\paragraph{Generator-centric methods.}
Most recent progress changes the generative model class: GraphDF uses a discrete flow
over molecular graphs~\cite{luo2021graphdf}, FREED an explorative reinforcement-learning policy over
fragments~\cite{yang2021freed}, GenMol a fragment-based discrete diffusion
model~\cite{genmol}, and DiGress a discrete graph-diffusion
process~\cite{vignac2023digress}. Our method introduces no generator parameters and adds a
cheap ranking layer over the candidate pool, making it complementary to these
advances, and in particular to GenMol, on which our largest gains appear.

\paragraph{Diversity and validity.}
The score gains come with an exploration-diversity trade-off. GFlowNets were introduced
precisely to sample diverse high-reward candidates rather than collapse to a
mode~\cite{bengio2021gflownet}, and preference-conditioned GFlowNets treat diversity as
a first-class objective in multi-objective molecular
optimization~\cite{zhu2023hngfn}; our deterministic selector sits at the
high-exploitation end of this spectrum, while the stochastic selector of
Section~\ref{subsec:selection} moves back toward diversity. A parallel line of
graph-generation work instead emphasizes chemical validity and structural
constraints, with hard constraints enforced throughout
sampling~\cite{madeira2024construct} and geometry-aware conditional
generation~\cite{huang20223dlinker}. This is complementary to our focus on oracle
allocation.

%======================================================================
\section{Conclusion}
\label{sec:conclusion}

We presented short-term graph memory, an oracle-budget-neutral plug-in that recycles oracle
feedback into an online graph surrogate to pre-screen future candidates. It improves
oracle-efficient molecular optimization across four backbones at a tight budget, and keeps
that advantage at the full budget for the two whose own search concentrates least. Our
analysis points to a compatibility question rather than a universal gain. The exploitation
bias that makes memory useful helps most where the base generator searches broadly, and helps
less once a generator's own search has narrowed.
The stochastic selector shows that this pressure is adjustable rather than fixed. Because
every backbone improves at the tight budget, early performance alone does not indicate which
will keep the advantage; the breadth of a generator's unguided search is the more promising
signal to test.

%======================================================================
\section{Limitations and Ethical Considerations}
\label{sec:limitations}
\textbf{Limitations.} Our evidence comes from an in-silico benchmark whose oracles are
predictive models rather than laboratory measurements. A real campaign may afford far fewer
evaluations than the thousand calls we treat as the tight budget, and a surrogate trained
within the run needs some feedback before it can rank anything at all; how far down the budget
scale this class of method stays useful is a question for validation against real assays
rather than for this benchmark. Our account of generator--memory compatibility is supported by
the observed score and diversity trajectories, although directly measuring surrogate ranking
quality within each proposal pool would provide a stronger test of the proposed explanation.
We hold the GraphGPS surrogate fixed across backbones to isolate generator--selector
interactions; comparing alternative surrogate architectures and uncertainty-aware selectors
remains future work. Finally, we study four backbone families, and broader generality remains
to be tested.
\\
\textbf{Ethical considerations.} This work studies oracle-efficient molecular optimization, a
tool intended to accelerate scientific discovery such as drug and materials design. All
experiments use public benchmark oracles and involve no human subjects, personal data, or
private information, so no informed consent or institutional review is required. The
objectives are the benchmark's own: drug-likeness, activity against published targets, and
similarity to approved drugs. Scope for misuse is bounded by what the method supplies, which
is neither the objective a search pursues nor any means of making a molecule. Its outputs are
2D structures scored by computational proxies, with no synthesis route and no experimental
validation, and are hypotheses for laboratory testing rather than results. We encourage
responsible use consistent with community norms for AI-assisted molecular design.

\bibliographystyle{ACM-Reference-Format}
\bibliography{refs}

%======================================================================
\appendix
\section{Implementation Details}
\label{app:impl}

The surrogate is a five-layer GraphGPS network~\cite{graphgps}. Each layer combines GINE
message passing~\cite{xu2018how,hu2020strategies} with four-head
attention~\cite{vaswani2017attention} at hidden dimension $300$, using Laplacian and
random-walk positional encodings. Atoms carry atomic number,
formal charge, chirality, hybridization, hydrogen count, valence, and degree; bonds carry
type and direction. Node embeddings are mean-pooled and mapped to a scalar by a linear head.

Online updates use Adam at learning rate $10^{-3}$, one gradient step per update on a
mini-batch of up to $256$ molecules sampled from the buffer, minimizing mean-squared error
against the recorded oracle scores. All network parameters are updated online, not just the
head, and each mini-batch is drawn uniformly from the accumulated buffer. The buffer is unbounded within a run and is discarded
when the budget is exhausted; nothing is carried across oracles.

Table~\ref{tab:setup} gives the per-backbone configuration. The proposal pool size $N$ and
forward count $k$ follow each backbone's own generation loop: the fragment generators produce
a pool per round and evaluate one molecule from it, while REINVENT and Graph-GA produce a
batch per round and evaluate half of it. Warm-up $w$ and update period $u$ are counted in
oracle calls and were set once per generation loop rather than tuned per oracle.

All runs used one NVIDIA RTX A6000 GPU each, on a host with two Intel Xeon Platinum 8352Y
CPUs ($64$ cores). Each run was given between $8$ and $24$ CPU threads, and several ran
concurrently on the same host. Oracle evaluation is CPU-bound, so throughput depends more on
the thread allocation than on the GPU.

\begin{table}[H]
\caption{Per-backbone configuration. $N$ is the candidate pool proposed each round, $k$ the
number forwarded to the oracle. Warm-up $w$ and update period $u$ are in oracle calls.}
\label{tab:setup}
\normalsize
\begin{tabular*}{\columnwidth}{@{\extracolsep{\fill}} l cc l cc @{}}
\toprule
Backbone & $N$ & $k$ & Selector & $w$ & $u$ \\
\midrule
GenMol      & 16 & 1  & deterministic & 50  & 10 \\
InVirtuoGen & 16 & 1  & deterministic & 50  & 10 \\
REINVENT    & 64 & 32 & stochastic    & 100 & 50 \\
Graph-GA    & 70 & 35 & deterministic & 100 & 50 \\
\bottomrule
\end{tabular*}
\end{table}

%======================================================================
\section{Molecules Found With and Without Memory}
\label{app:mols}

\begin{figure*}[t]
\centering
\includegraphics[width=\textwidth]{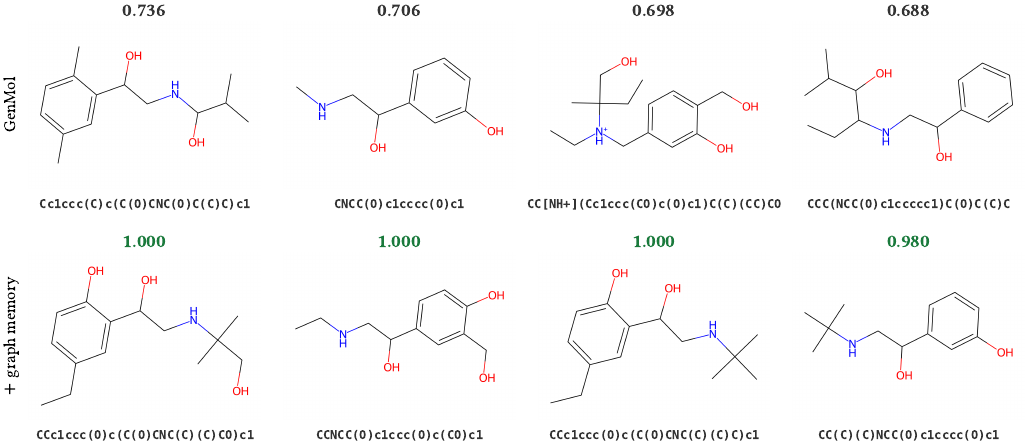}
\caption{Four highest-scoring molecules found by GenMol on albuterol similarity, with and
without graph memory, under the same $10{,}000$-call budget. Each panel gives the oracle
score and the SMILES string.}
\Description{A two-by-four grid of molecular structure diagrams. The top row shows the four
highest-scoring molecules found by the base GenMol generator, scoring between 0.688 and
0.736. The bottom row shows those found with graph memory, three of which score 1.000 and one
0.980. Each structure is captioned with its oracle score and SMILES string. The bottom-row
structures share the phenol, secondary alcohol and bulky amine of the albuterol target, which
the top-row structures lack.}
\label{fig:mols}
\end{figure*}

Figure~\ref{fig:mols} shows what the score gap looks like structurally on albuterol
similarity, one of the oracles where GenMol improves most with graph memory. The base
generator's best molecules carry a
$\beta$-hydroxy-ethylamine motif but miss the target's substitution pattern, and one of them
is a quaternary ammonium the target does not have. With memory, all four best molecules pair
that motif with the phenol and the bulky amine of albuterol, and three of them reach the
maximum score. Graph memory does not modify the proposal operator; here it reallocates the budget
toward generated candidates closer to the target.

%======================================================================
\section{Hard Selection and Policy-Gradient Collapse}
\label{app:reinvent}

In REINVENT the candidates that survive selection are also the batch on which the policy
gradient is computed. A hard filter therefore feeds the selector's bias back through the
generator's parameters, where it is recursively amplified. The policy concentrates on the
modes the surrogate prefers and begins emitting duplicates. Duplicate molecules are served
from the oracle's cache and consume no budget, so the oracle counter stops advancing and the
run stalls well before the budget is spent. Table~\ref{tab:hardsoft} reports the oracle calls
completed before termination; a run was stopped once its oracle count had stayed unchanged
for $30$ minutes on the hardware of Appendix~\ref{app:impl}. The threshold has only to tell a
frozen counter from a slow one: a stalled run draws every proposal from the cache, so its
count does not advance at all rather than advancing slowly.

Table~\ref{tab:hardsoft} reports every oracle for which we attempted a hard-filter run. Three of the
six stall, one as early as $600$ calls. The same oracles with the stochastic selector all
complete the full budget. Graph-GA filters the same fraction of its batch, $35$ of $70$, and
with a hard filter completes all $23$ oracles without stalling. Its selector output also
propagates forward, through the population that seeds the next generation, but fixed mutation,
crossover, and stochastic mating keep injecting variation, and in our runs that was enough to
prevent the same collapse. Under the tested configurations, hard selection stalled REINVENT
but not Graph-GA.

This is why the REINVENT runs in Table~\ref{tab:main-final} use the stochastic selector, with
$T{=}1$ throughout.

\begin{table}[H]
\caption{Oracle calls completed out of $10{,}000$ by REINVENT, on every oracle we ran
with a hard filter.}
\label{tab:hardsoft}
\normalsize
\begin{tabular*}{\columnwidth}{@{\extracolsep{\fill}} l cc @{}}
\toprule
Oracle & Hard filter & Stochastic \\
\midrule
isomers c7h8n2o2 & 600    & 10{,}000 \\
drd2             & 2{,}100 & 10{,}000 \\
jnk3             & 9{,}800 & 10{,}000 \\
deco hop         & 10{,}000 & 10{,}000 \\
median1          & 10{,}000 & 10{,}000 \\
scaffold hop     & 10{,}000 & 10{,}000 \\
\bottomrule
\end{tabular*}
\end{table}

%======================================================================
\section{The valsartan SMARTS Oracle}
\label{app:valsartan}

The valsartan oracle rewards molecules that match a fixed SMARTS substructure subject to
several physicochemical constraints. Its top score is near-binary in practice: a run either
discovers the substructure and scores near $1$, or never finds it and scores near $0$, with
almost nothing in between. Table~\ref{tab:valsartan} shows the final top-10 score on each
backbone. Two generators never find the pattern and score $0$, GenMol finds it with memory but
not without, and REINVENT finds it in some runs but not others, so on this oracle the recorded
value reflects whether a run happened to encounter the substructure. The other oracles we
examined did not show the same near-binary run-to-run behavior.

Because this instability was identified after the initial runs, we report valsartan separately
and examine its effect on the aggregate. Including it adds GenMol's single largest per-oracle
gain, from $0.05$ at base to $0.90$ with memory, while leaving the other backbones' aggregates
essentially unchanged. The numerical aggregates change, but the qualitative conclusions do
not.
\begin{table}[H]
\caption{Final top-10 score on the valsartan oracle.\label{tab:valsartan}}
\normalsize
\begin{tabular*}{\columnwidth}{@{\extracolsep{\fill}} l cc @{}}
\toprule
Backbone & Base & +Graph Mem \\
\midrule
GenMol       & 0.051 & \textbf{0.902} \\
InVirtuoGen  & 0.000 & 0.000 \\
REINVENT     & 0.988 & 0.987 \\
Graph-GA     & 0.000 & 0.000 \\
\bottomrule
\end{tabular*}
\end{table}
%======================================================================
\begin{table*}[!b]
\caption{\textbf{Final top-10 score per oracle at a $1{,}000$-call budget}, base generator vs.\ short-term graph memory. The value after $\pm$ is the standard deviation over three random seeds; it averages $0.02$ and never exceeds $0.10$. \label{tab:per-oracle-1k}}
\fontsize{9}{11}\selectfont
\renewcommand{\arraystretch}{1.25}
\begin{tabular*}{\textwidth}{@{\extracolsep{\fill}} l cc cc cc cc @{}}
\toprule
 & \multicolumn{2}{c}{GenMol} & \multicolumn{2}{c}{InVirtuoGen} & \multicolumn{2}{c}{REINVENT} & \multicolumn{2}{c}{Graph-GA} \\
\cmidrule(lr){2-3}\cmidrule(lr){4-5}\cmidrule(lr){6-7}\cmidrule(lr){8-9}
Oracle & Base & +Mem & Base & +Mem & Base & +Mem & Base & +Mem \\
\midrule
albuterol similarity & 0.567\sd{0.01} & \textbf{0.673\sd{0.01}} & 0.566\sd{0.01} & \textbf{0.574\sd{0.02}} & 0.583\sd{0.03} & \textbf{0.612\sd{0.02}} & 0.508\sd{0.06} & \textbf{0.690\sd{0.03}} \\
amlodipine mpo & 0.571\sd{0.01} & \textbf{0.599\sd{0.01}} & 0.536\sd{0.01} & \textbf{0.537\sd{0.01}} & 0.496\sd{0.01} & \textbf{0.508\sd{0.01}} & \textbf{0.539\sd{0.01}} & 0.531\sd{0.01} \\
celecoxib rediscovery & 0.456\sd{0.01} & \textbf{0.529\sd{0.01}} & 0.413\sd{0.03} & \textbf{0.453\sd{0.03}} & 0.426\sd{0.03} & \textbf{0.454\sd{0.03}} & 0.385\sd{0.01} & \textbf{0.457\sd{0.02}} \\
deco hop & 0.641\sd{0.03} & \textbf{0.648\sd{0.10}} & 0.612\sd{0.00} & 0.612\sd{0.01} & \textbf{0.591\sd{0.01}} & 0.589\sd{0.01} & \textbf{0.607\sd{0.01}} & 0.582\sd{0.00} \\
drd2 & 0.984\sd{0.00} & \textbf{0.999\sd{0.00}} & 0.993\sd{0.01} & \textbf{1.000\sd{0.00}} & 0.975\sd{0.01} & \textbf{0.999\sd{0.00}} & 0.999\sd{0.01} & 0.999\sd{0.01} \\
fexofenadine mpo & 0.731\sd{0.01} & \textbf{0.750\sd{0.00}} & \textbf{0.702\sd{0.02}} & 0.698\sd{0.01} & \textbf{0.695\sd{0.01}} & 0.679\sd{0.00} & 0.699\sd{0.00} & \textbf{0.739\sd{0.03}} \\
gsk3b & 0.673\sd{0.01} & \textbf{0.820\sd{0.02}} & 0.572\sd{0.09} & \textbf{0.932\sd{0.06}} & \textbf{0.773\sd{0.08}} & 0.681\sd{0.05} & \textbf{0.590\sd{0.03}} & 0.569\sd{0.02} \\
isomers c7h8n2o2 & 0.485\sd{0.05} & \textbf{0.865\sd{0.02}} & 0.597\sd{0.05} & \textbf{0.629\sd{0.02}} & 0.879\sd{0.04} & \textbf{0.983\sd{0.04}} & \textbf{0.846\sd{0.02}} & 0.840\sd{0.07} \\
isomers c9h10n2o2pf2cl & 0.553\sd{0.04} & \textbf{0.657\sd{0.03}} & 0.672\sd{0.10} & \textbf{0.771\sd{0.03}} & 0.696\sd{0.05} & \textbf{0.699\sd{0.07}} & 0.723\sd{0.04} & \textbf{0.763\sd{0.07}} \\
jnk3 & 0.492\sd{0.01} & \textbf{0.673\sd{0.01}} & 0.484\sd{0.08} & \textbf{0.575\sd{0.07}} & 0.205\sd{0.02} & \textbf{0.561\sd{0.10}} & \textbf{0.244\sd{0.03}} & 0.219\sd{0.06} \\
median1 & 0.236\sd{0.00} & \textbf{0.274\sd{0.01}} & \textbf{0.243\sd{0.01}} & 0.226\sd{0.01} & 0.236\sd{0.02} & \textbf{0.275\sd{0.01}} & 0.222\sd{0.01} & \textbf{0.264\sd{0.00}} \\
median2 & 0.244\sd{0.00} & \textbf{0.272\sd{0.01}} & 0.187\sd{0.02} & \textbf{0.221\sd{0.01}} & 0.194\sd{0.00} & \textbf{0.209\sd{0.01}} & 0.211\sd{0.01} & \textbf{0.224\sd{0.02}} \\
mestranol similarity & 0.438\sd{0.01} & \textbf{0.516\sd{0.00}} & 0.408\sd{0.02} & \textbf{0.412\sd{0.02}} & 0.354\sd{0.04} & \textbf{0.475\sd{0.02}} & 0.399\sd{0.03} & \textbf{0.462\sd{0.05}} \\
osimertinib mpo & 0.815\sd{0.00} & \textbf{0.823\sd{0.00}} & \textbf{0.783\sd{0.00}} & 0.779\sd{0.01} & 0.776\sd{0.01} & \textbf{0.788\sd{0.01}} & \textbf{0.793\sd{0.00}} & 0.788\sd{0.01} \\
perindopril mpo & 0.517\sd{0.01} & \textbf{0.523\sd{0.01}} & 0.454\sd{0.01} & \textbf{0.493\sd{0.02}} & 0.417\sd{0.01} & \textbf{0.449\sd{0.00}} & 0.471\sd{0.01} & \textbf{0.480\sd{0.01}} \\
qed & 0.931\sd{0.01} & \textbf{0.944\sd{0.00}} & 0.936\sd{0.00} & \textbf{0.937\sd{0.00}} & 0.940\sd{0.00} & \textbf{0.941\sd{0.00}} & 0.939\sd{0.00} & \textbf{0.942\sd{0.00}} \\
ranolazine mpo & 0.686\sd{0.01} & \textbf{0.757\sd{0.01}} & 0.687\sd{0.02} & \textbf{0.706\sd{0.02}} & 0.639\sd{0.01} & \textbf{0.648\sd{0.01}} & 0.646\sd{0.02} & \textbf{0.720\sd{0.03}} \\
scaffold hop & 0.506\sd{0.00} & \textbf{0.512\sd{0.01}} & 0.486\sd{0.01} & \textbf{0.487\sd{0.00}} & 0.467\sd{0.00} & \textbf{0.473\sd{0.00}} & \textbf{0.473\sd{0.01}} & 0.471\sd{0.01} \\
sitagliptin mpo & 0.228\sd{0.02} & \textbf{0.369\sd{0.01}} & \textbf{0.338\sd{0.04}} & 0.315\sd{0.02} & 0.310\sd{0.00} & \textbf{0.379\sd{0.04}} & 0.332\sd{0.02} & \textbf{0.436\sd{0.06}} \\
thiothixene rediscovery & 0.397\sd{0.01} & \textbf{0.458\sd{0.01}} & 0.342\sd{0.02} & \textbf{0.419\sd{0.03}} & 0.326\sd{0.02} & \textbf{0.394\sd{0.03}} & 0.323\sd{0.01} & \textbf{0.333\sd{0.01}} \\
troglitazone rediscovery & 0.317\sd{0.01} & \textbf{0.362\sd{0.03}} & 0.266\sd{0.01} & \textbf{0.298\sd{0.01}} & 0.260\sd{0.00} & \textbf{0.301\sd{0.02}} & 0.287\sd{0.01} & \textbf{0.315\sd{0.01}} \\
zaleplon mpo & 0.460\sd{0.01} & \textbf{0.466\sd{0.01}} & \textbf{0.436\sd{0.00}} & 0.432\sd{0.01} & 0.448\sd{0.00} & \textbf{0.468\sd{0.00}} & 0.432\sd{0.01} & \textbf{0.439\sd{0.01}} \\
\midrule
\textbf{Mean} & 0.542 & \textbf{0.613} & 0.532 & \textbf{0.569} & 0.531 & \textbf{0.571} & 0.530 & \textbf{0.557} \\
\textbf{W/T/L} & \multicolumn{2}{c}{22/0/0} & \multicolumn{2}{c}{16/1/5} & \multicolumn{2}{c}{19/0/3} & \multicolumn{2}{c}{14/1/7} \\
\textbf{Wilcoxon $p$} & \multicolumn{2}{c}{$\mathbf{2{\times}10^{-7}}$} & \multicolumn{2}{c}{0.002} & \multicolumn{2}{c}{$\mathbf{5{\times}10^{-4}}$} & \multicolumn{2}{c}{0.009} \\
\bottomrule
\end{tabular*}
\end{table*}

\section{Augmented Memory}
\label{app:augmem}
\begin{algorithm}[htbp]
\caption{Schematic of Augmented Memory~\cite{augmem}}
\label{alg:augmem}
\begin{algorithmic}[1]
\Require generator $G_\theta$, oracle $f$, budget $B$, replay buffer $\mathcal{R}$, augmentation rounds $A$
\State $\mathcal{M} \gets \emptyset$;\ \ $\mathcal{R} \gets \emptyset$
\While{$|\mathcal{M}| < B$}
  \State $X \gets G_\theta.\textsc{sample}()$
  \State $y \gets f(X)$;\ \ $\mathcal{M} \gets \mathcal{M} \cup \{(X,y)\}$
  \State $\mathcal{R} \gets \textsc{keep-top}(\mathcal{R} \cup \{(X,y)\})$
  \State $\theta \gets \textsc{policy-update}(\theta,\ X,\ y)$
  \For{$a = 1 \dots A$} \label{line:augreplay}
    \State $X' \gets \textsc{randomized-SMILES}(\mathcal{R})$ \Comment{no oracle calls}
    \State $\theta \gets \textsc{policy-update}(\theta,\ X',\ \mathcal{R}.\text{scores})$
  \EndFor
\EndWhile
\State \Return top molecules in $\mathcal{M}$
\end{algorithmic}
\end{algorithm}

Algorithm~\ref{alg:augmem} sketches Augmented Memory~\cite{augmem} in the same notation as
Algorithm~\ref{alg:graphmem}, omitting the likelihood and training details specific to that
method. Line~\ref{line:augreplay} reuses buffered high scorers to shape further generation
without spending oracle calls, which the original does by policy gradient. Because a frozen
fragment generator has no policy to update, our adaptation replaces the replayed policy
updates with remasking and regeneration around the buffered molecules.

The buffer keeps the top $100$ unique molecules with a non-zero score. After warm-up, each
round remasks a buffered molecule with probability $0.5$, drawn in proportion to its score,
and otherwise falls back to GenMol's ordinary fragment recombination. Following the reference
implementation, the buffer is cleared when its upper half collapses onto a single sub-optimal
score; we apply this rule with the published threshold on every oracle. This run and the
graph-memory run share the fragment pool, the candidate-pool size, the oracle budget, the
oracle-cache accounting, and every other GenMol setting. The two differ in whether past
feedback is reused through reseeding or through external ranking.

%======================================================================
\section{Per-Oracle Results}
\label{app:auc}

Tables~\ref{tab:per-oracle-1k} and~\ref{tab:per-oracle} report final scores by oracle
for all four backbones at the tight and full budgets; Tables~\ref{tab:per-oracle-auc1k}
and~\ref{tab:main-auc} report climbing speed. The last rows give the wins/ties/losses
and one-sided Wilcoxon $p$-values for memory exceeding base across the 22 paired
oracles. Entries are rounded to three decimals, and wins, ties, and losses are counted
at that same precision, so a tie means the two values agree to three decimals. The
Wilcoxon test uses the unrounded scores and resolves zero differences by the Pratt
method~\cite{pratt1959remarks}, which ranks them with the rest and then drops their ranks from the statistic.

\begin{table*}[t]
\caption{Per-oracle top-10 AUC at a $1{,}000$-call budget, base generator vs.\ short-term graph memory.\label{tab:per-oracle-auc1k}}
\small
\renewcommand{\arraystretch}{0.92}
\begin{tabular*}{\textwidth}{@{\extracolsep{\fill}} l cc cc cc cc @{}}
\toprule
 & \multicolumn{2}{c}{GenMol} & \multicolumn{2}{c}{InVirtuoGen} & \multicolumn{2}{c}{REINVENT} & \multicolumn{2}{c}{Graph-GA} \\
\cmidrule(lr){2-3}\cmidrule(lr){4-5}\cmidrule(lr){6-7}\cmidrule(lr){8-9}
Oracle & Base & +Mem & Base & +Mem & Base & +Mem & Base & +Mem \\
\midrule
albuterol similarity & 0.561 & \textbf{0.663} & \textbf{0.482} & 0.454 & 0.465 & \textbf{0.469} & 0.407 & \textbf{0.479} \\
amlodipine mpo & 0.566 & \textbf{0.593} & 0.475 & \textbf{0.483} & 0.452 & \textbf{0.460} & \textbf{0.480} & 0.468 \\
celecoxib rediscovery & 0.450 & \textbf{0.521} & 0.311 & \textbf{0.349} & 0.334 & 0.334 & 0.290 & \textbf{0.345} \\
deco hop & 0.635 & \textbf{0.642} & 0.559 & \textbf{0.560} & \textbf{0.552} & 0.548 & \textbf{0.558} & 0.547 \\
drd2 & 0.976 & \textbf{0.992} & 0.921 & \textbf{0.923} & 0.543 & \textbf{0.732} & 0.751 & \textbf{0.765} \\
fexofenadine mpo & 0.727 & \textbf{0.744} & \textbf{0.641} & 0.632 & \textbf{0.626} & 0.613 & 0.625 & \textbf{0.648} \\
gsk3b & 0.665 & \textbf{0.806} & 0.451 & \textbf{0.722} & \textbf{0.450} & 0.423 & 0.442 & \textbf{0.452} \\
isomers c7h8n2o2 & 0.478 & \textbf{0.844} & \textbf{0.418} & 0.342 & 0.476 & \textbf{0.666} & 0.681 & \textbf{0.697} \\
isomers c9h10n2o2pf2cl & 0.536 & \textbf{0.644} & 0.500 & \textbf{0.653} & 0.541 & \textbf{0.549} & 0.557 & \textbf{0.597} \\
jnk3 & 0.485 & \textbf{0.662} & 0.318 & \textbf{0.344} & 0.147 & \textbf{0.297} & \textbf{0.171} & 0.162 \\
median1 & 0.234 & \textbf{0.270} & \textbf{0.190} & 0.175 & 0.190 & \textbf{0.207} & 0.173 & \textbf{0.190} \\
median2 & 0.242 & \textbf{0.269} & 0.160 & \textbf{0.185} & 0.160 & \textbf{0.173} & 0.173 & \textbf{0.180} \\
mestranol similarity & 0.433 & \textbf{0.509} & 0.337 & \textbf{0.348} & 0.302 & \textbf{0.371} & 0.316 & \textbf{0.339} \\
osimertinib mpo & 0.811 & \textbf{0.818} & 0.720 & \textbf{0.721} & 0.716 & \textbf{0.721} & 0.718 & \textbf{0.729} \\
perindopril mpo & 0.512 & \textbf{0.520} & 0.401 & \textbf{0.444} & 0.379 & \textbf{0.395} & 0.410 & \textbf{0.411} \\
qed & 0.924 & \textbf{0.938} & 0.875 & \textbf{0.879} & \textbf{0.884} & 0.883 & 0.887 & \textbf{0.888} \\
ranolazine mpo & 0.681 & \textbf{0.750} & 0.568 & \textbf{0.600} & 0.497 & \textbf{0.514} & 0.499 & \textbf{0.543} \\
scaffold hop & 0.503 & \textbf{0.509} & \textbf{0.445} & 0.438 & 0.430 & 0.430 & 0.434 & \textbf{0.436} \\
sitagliptin mpo & 0.219 & \textbf{0.358} & \textbf{0.248} & 0.243 & 0.192 & \textbf{0.264} & 0.221 & \textbf{0.272} \\
thiothixene rediscovery & 0.394 & \textbf{0.453} & 0.278 & \textbf{0.327} & 0.274 & \textbf{0.292} & 0.265 & \textbf{0.284} \\
troglitazone rediscovery & 0.314 & \textbf{0.358} & 0.219 & \textbf{0.241} & 0.219 & \textbf{0.234} & 0.229 & \textbf{0.240} \\
zaleplon mpo & 0.456 & \textbf{0.460} & \textbf{0.379} & 0.364 & 0.385 & \textbf{0.402} & 0.375 & \textbf{0.385} \\
\midrule
\textbf{Mean} & 0.537 & \textbf{0.606} & 0.450 & \textbf{0.474} & 0.419 & \textbf{0.453} & 0.439 & \textbf{0.457} \\
\textbf{W/T/L} & \multicolumn{2}{c}{22/0/0} & \multicolumn{2}{c}{15/0/7} & \multicolumn{2}{c}{16/2/4} & \multicolumn{2}{c}{19/0/3} \\
\textbf{Wilcoxon $p$} & \multicolumn{2}{c}{$\mathbf{2{\times}10^{-7}}$} & \multicolumn{2}{c}{0.049} & \multicolumn{2}{c}{$\mathbf{1{\times}10^{-3}}$} & \multicolumn{2}{c}{$\mathbf{3{\times}10^{-4}}$} \\
\bottomrule
\end{tabular*}
\end{table*}

\begin{table*}[t]
\caption{\textbf{Final top-10 score per oracle at a $10{,}000$-call budget}, base generator vs.\ short-term graph memory.\label{tab:per-oracle}}
\small
\renewcommand{\arraystretch}{0.92}
\begin{tabular*}{\textwidth}{@{\extracolsep{\fill}} l cc cc cc cc @{}}
\toprule
 & \multicolumn{2}{c}{GenMol} & \multicolumn{2}{c}{InVirtuoGen} & \multicolumn{2}{c}{REINVENT} & \multicolumn{2}{c}{Graph-GA} \\
\cmidrule(lr){2-3}\cmidrule(lr){4-5}\cmidrule(lr){6-7}\cmidrule(lr){8-9}
Oracle & Base & +Mem & Base & +Mem & Base & +Mem & Base & +Mem \\
\midrule
albuterol similarity & 0.686 & \textbf{0.996} & 0.896 & \textbf{1.000} & 0.997 & \textbf{1.000} & 1.000 & 1.000 \\
amlodipine mpo & 0.612 & \textbf{0.706} & \textbf{0.679} & 0.656 & \textbf{0.806} & 0.741 & 0.766 & \textbf{0.818} \\
celecoxib rediscovery & 0.531 & \textbf{0.785} & 0.733 & \textbf{0.892} & 0.882 & 0.882 & 0.558 & \textbf{0.832} \\
deco hop & 0.857 & \textbf{0.944} & 0.640 & \textbf{0.679} & \textbf{0.707} & 0.678 & \textbf{0.925} & 0.716 \\
drd2 & 1.000 & 1.000 & 1.000 & 1.000 & 1.000 & 1.000 & 1.000 & 1.000 \\
fexofenadine mpo & 0.754 & \textbf{0.841} & \textbf{0.835} & 0.833 & \textbf{0.890} & 0.887 & 0.847 & \textbf{0.880} \\
gsk3b & 0.830 & \textbf{0.970} & 0.964 & \textbf{1.000} & 1.000 & 1.000 & \textbf{0.972} & 0.881 \\
isomers c7h8n2o2 & 0.941 & \textbf{1.000} & 0.816 & \textbf{0.936} & 1.000 & 1.000 & 1.000 & 1.000 \\
isomers c9h10n2o2pf2cl & 0.816 & \textbf{0.877} & 0.801 & \textbf{0.874} & 0.911 & \textbf{0.916} & \textbf{0.888} & 0.882 \\
jnk3 & 0.684 & \textbf{0.938} & 0.811 & \textbf{0.902} & \textbf{0.984} & 0.750 & \textbf{0.896} & 0.440 \\
median1 & 0.322 & \textbf{0.399} & \textbf{0.332} & 0.328 & 0.450 & 0.450 & \textbf{0.324} & 0.322 \\
median2 & 0.295 & \textbf{0.387} & 0.285 & \textbf{0.333} & \textbf{0.344} & 0.317 & 0.319 & \textbf{0.324} \\
mestranol similarity & 0.650 & \textbf{1.000} & 0.633 & \textbf{0.635} & \textbf{0.802} & 0.700 & \textbf{0.724} & 0.647 \\
osimertinib mpo & 0.828 & \textbf{0.864} & 0.838 & \textbf{0.849} & 0.912 & \textbf{0.926} & \textbf{0.862} & 0.854 \\
perindopril mpo & 0.546 & \textbf{0.643} & \textbf{0.603} & 0.582 & 0.609 & \textbf{0.626} & 0.578 & \textbf{0.579} \\
qed & 0.947 & \textbf{0.948} & 0.947 & \textbf{0.948} & 0.948 & 0.948 & 0.948 & 0.948 \\
ranolazine mpo & 0.719 & \textbf{0.799} & 0.822 & \textbf{0.826} & 0.862 & \textbf{0.878} & 0.814 & \textbf{0.832} \\
scaffold hop & 0.534 & \textbf{0.597} & \textbf{0.570} & 0.556 & 0.613 & \textbf{0.635} & \textbf{0.639} & 0.559 \\
sitagliptin mpo & 0.376 & \textbf{0.486} & 0.491 & \textbf{0.570} & 0.544 & \textbf{0.580} & 0.697 & \textbf{0.763} \\
thiothixene rediscovery & 0.468 & \textbf{0.771} & 0.514 & \textbf{0.570} & \textbf{0.694} & 0.581 & 0.640 & \textbf{0.651} \\
troglitazone rediscovery & 0.379 & \textbf{0.735} & 0.368 & \textbf{0.560} & 0.631 & \textbf{0.676} & 0.644 & \textbf{0.668} \\
zaleplon mpo & 0.489 & \textbf{0.564} & 0.532 & \textbf{0.536} & 0.578 & \textbf{0.585} & 0.523 & \textbf{0.540} \\
\midrule
\textbf{Mean} & 0.648 & \textbf{0.784} & 0.687 & \textbf{0.730} & \textbf{0.780} & 0.762 & \textbf{0.753} & 0.733 \\
\textbf{W/T/L} & \multicolumn{2}{c}{21/1/0} & \multicolumn{2}{c}{16/1/5} & \multicolumn{2}{c}{9/6/7} & \multicolumn{2}{c}{10/4/8} \\
\textbf{Wilcoxon $p$} & \multicolumn{2}{c}{$\mathbf{2{\times}10^{-7}}$} & \multicolumn{2}{c}{0.002} & \multicolumn{2}{c}{0.56} & \multicolumn{2}{c}{0.41} \\
\bottomrule
\end{tabular*}
\end{table*}

\begin{table*}[t]
\caption{Per-oracle top-10 AUC at a $10{,}000$-call budget, base generator vs.\ short-term graph memory.\label{tab:main-auc}}
\small
\renewcommand{\arraystretch}{0.92}
\begin{tabular*}{\textwidth}{@{\extracolsep{\fill}} l cc cc cc cc @{}}
\toprule
 & \multicolumn{2}{c}{GenMol} & \multicolumn{2}{c}{InVirtuoGen} & \multicolumn{2}{c}{REINVENT} & \multicolumn{2}{c}{Graph-GA} \\
\cmidrule(lr){2-3}\cmidrule(lr){4-5}\cmidrule(lr){6-7}\cmidrule(lr){8-9}
Oracle & Base & +Mem & Base & +Mem & Base & +Mem & Base & +Mem \\
\midrule
albuterol similarity & 0.642 & \textbf{0.840} & 0.760 & \textbf{0.895} & \textbf{0.906} & 0.896 & 0.863 & \textbf{0.915} \\
amlodipine mpo & 0.589 & \textbf{0.657} & \textbf{0.612} & 0.607 & \textbf{0.674} & 0.659 & 0.682 & \textbf{0.684} \\
celecoxib rediscovery & 0.484 & \textbf{0.644} & 0.565 & \textbf{0.749} & 0.718 & \textbf{0.782} & 0.499 & \textbf{0.549} \\
deco hop & 0.749 & \textbf{0.877} & 0.628 & \textbf{0.643} & 0.641 & 0.641 & \textbf{0.759} & 0.626 \\
drd2 & 0.987 & \textbf{0.993} & 0.992 & 0.992 & 0.954 & \textbf{0.958} & 0.975 & \textbf{0.976} \\
fexofenadine mpo & 0.739 & \textbf{0.796} & \textbf{0.784} & 0.773 & 0.788 & \textbf{0.796} & 0.784 & \textbf{0.826} \\
gsk3b & 0.757 & \textbf{0.921} & 0.823 & \textbf{0.966} & 0.925 & \textbf{0.929} & \textbf{0.813} & 0.774 \\
isomers c7h8n2o2 & 0.812 & \textbf{0.950} & 0.707 & \textbf{0.802} & 0.944 & \textbf{0.960} & \textbf{0.938} & 0.932 \\
isomers c9h10n2o2pf2cl & 0.734 & \textbf{0.812} & 0.743 & \textbf{0.828} & 0.848 & \textbf{0.877} & 0.835 & \textbf{0.845} \\
jnk3 & 0.591 & \textbf{0.840} & 0.692 & \textbf{0.793} & \textbf{0.738} & 0.662 & \textbf{0.540} & 0.363 \\
median1 & 0.293 & \textbf{0.349} & \textbf{0.303} & 0.291 & 0.387 & \textbf{0.407} & 0.287 & \textbf{0.296} \\
median2 & 0.275 & \textbf{0.343} & 0.251 & \textbf{0.283} & 0.280 & \textbf{0.282} & 0.283 & \textbf{0.290} \\
mestranol similarity & 0.547 & \textbf{0.886} & \textbf{0.540} & 0.538 & 0.628 & \textbf{0.643} & 0.543 & \textbf{0.555} \\
osimertinib mpo & 0.816 & \textbf{0.841} & 0.812 & \textbf{0.818} & 0.834 & \textbf{0.856} & \textbf{0.832} & 0.816 \\
perindopril mpo & 0.525 & \textbf{0.569} & 0.530 & \textbf{0.538} & 0.524 & \textbf{0.549} & 0.523 & \textbf{0.536} \\
qed & 0.938 & \textbf{0.942} & 0.938 & \textbf{0.939} & 0.941 & 0.941 & 0.941 & 0.941 \\
ranolazine mpo & 0.703 & \textbf{0.769} & 0.768 & \textbf{0.784} & 0.774 & \textbf{0.788} & 0.742 & \textbf{0.770} \\
scaffold hop & 0.518 & \textbf{0.559} & \textbf{0.523} & 0.521 & 0.556 & \textbf{0.560} & \textbf{0.561} & 0.524 \\
sitagliptin mpo & 0.320 & \textbf{0.433} & 0.412 & \textbf{0.467} & 0.455 & \textbf{0.505} & 0.509 & \textbf{0.626} \\
thiothixene rediscovery & 0.425 & \textbf{0.592} & 0.448 & \textbf{0.510} & \textbf{0.572} & 0.530 & 0.493 & \textbf{0.533} \\
troglitazone rediscovery & 0.345 & \textbf{0.540} & 0.337 & \textbf{0.464} & 0.483 & \textbf{0.541} & 0.457 & \textbf{0.487} \\
zaleplon mpo & 0.475 & \textbf{0.524} & \textbf{0.493} & 0.484 & 0.524 & \textbf{0.542} & 0.477 & \textbf{0.497} \\
\midrule
\textbf{Mean} & 0.603 & \textbf{0.712} & 0.621 & \textbf{0.668} & 0.686 & \textbf{0.696} & 0.652 & \textbf{0.653} \\
\textbf{W/T/L} & \multicolumn{2}{c}{22/0/0} & \multicolumn{2}{c}{15/1/6} & \multicolumn{2}{c}{16/2/4} & \multicolumn{2}{c}{15/1/6} \\
\textbf{Wilcoxon $p$} & \multicolumn{2}{c}{$\mathbf{2{\times}10^{-7}}$} & \multicolumn{2}{c}{0.002} & \multicolumn{2}{c}{0.015} & \multicolumn{2}{c}{0.099} \\
\bottomrule
\end{tabular*}
\end{table*}

\end{document}